\documentclass[10pt,twocolumn,a4paper]{article}
\usepackage[a4paper, total={7in, 10.4in}]{geometry}
\usepackage[numbers,sort]{natbib}
\usepackage[font=small]{caption}
\usepackage[utf8]{inputenc}
\usepackage{rotating}
\usepackage{amssymb}
\usepackage{csquotes}
\usepackage{graphicx}
\usepackage{times}
\usepackage{float}
\usepackage{ifpdf}
\usepackage{acro}
\usepackage[caption=false]{subfig}
\ifpdf
\fi
\newcommand{\cites}[1]{\citeauthor{#1}'s\ \citep{#1}}
\newcommand{\rot}[2]{\begin{turn}{#1}\textbf{#2}\end{turn}}
\DeclareAcronym{ie}{short = i.e., long = that is}
\DeclareAcronym{eg}{short = e.g., long = Example}
\DeclareAcronym{aka}{short=a.k.a.,long=also known as}
\DeclareAcronym{slc}{short=SLC,long=Single-Label Classification}
\DeclareAcronym{mlc}{short=MLC,long=Multi-Label Classification}
\DeclareAcronym{cam}{short = CAM, long = Class Activation Map}
\DeclareAcronym{roi}{short = RoI, long = Region of Interest}
\DeclareAcronym{rpn}{short=RPN,long=Region Proposal Network}
\DeclareAcronym{rcnn}{short=R-CNN,long=Region-based Convolutional Neural Network}
\DeclareAcronym{spp}{short=SPP,long=Spatial Pyramid Pooling}
\DeclareAcronym{cnn}{short=CNN,long=Convolutional Neural Network}
\DeclareAcronym{drnet}{short = DR-Net, long = Deep Relational Network}
\DeclareAcronym{vgg}{short=VGG,long=Visual Geometry Group}
\DeclareAcronym{vgg16}{short=VGG-16,long=Visual Geometry Group 16-Layer}
\DeclareAcronym{vgg19}{short=VGG-19,long=Visual Geometry Group 19-layer}
\DeclareAcronym{fastrcnn}{short=Fast R-CNN,long=Fast Region-based Convolutional Neural Network}
\DeclareAcronym{fasterrcnn}{short=Faster R-CNN,long=Faster Region-based Convolutional Neural Network}
\DeclareAcronym{maskrcnn}{short=Mask R-CNN,long=Mask Region-based Convolutional Neural Network}
\DeclareAcronym{vg}{short=VG,long=Visual Genome}
\DeclareAcronym{vrd}{short=VRD,long=Visual Relationship Detection}
\DeclareAcronym{vqa}{short = VQA, long = Visual Question Answering}
\DeclareAcronym{mscoco}{short = MS COCO, long = Microsoft Common Objects in COntext}
\DeclareAcronym{sgd}{short=SGD,long=Stochastic Gradient Descent}
\DeclareAcronym{rgb}{short = RGB, long = {Red, Green and Blue}}
\DeclareAcronym{pov}{short = POV, long = Point-of-View}
\DeclareAcronym{api}{short = API, long = Application Programming Interface}
\DeclareAcronym{map}{short=MAP,long=Maximum A Posteriori}
\DeclareAcronym{gpu}{short=GPU,long=Graphics Processing Unit}
\DeclareAcronym{lstm}{short=LSTM,long=Long-Short Term Memory}

\newcommand{\mycaptions}{\centering}

\begin{document}
	
	\title{\textbf{Optimising the Input Image to Improve Visual Relationship Detection}}
	
	\author{
		Noel Mizzi\\
		Department of Computer Science \\
		University of Malta, Malta \\
		\and	
		Adrian Muscat\\
		Department of Computer Science \\
		University of Malta, Malta
	}
	\date{}
	\maketitle
	
	\begin{abstract}
		Visual Relationship Detection is defined as, given an image composed of a subject and an object, the correct relation is predicted. To improve the visual part of this difficult problem, ten preprocessing methods were tested to determine whether the widely-used Union method yields the optimal results. Therefore, focusing solely on predicate prediction, no object detection and linguistic knowledge were used to prevent them from affecting the comparison results. Once fine-tuned, the Visual Geometry Group models were evaluated using Recall@1, per-predicate recall, activation maximisations, class activation maps, and error analysis. From this research it was found that using preprocessing methods such as the Union-Without-Background-and-with-Binary-mask (Union-WB-and-B) method yields significantly better results than the widely-used Union method since, as designed, it enables the Convolutional Neural Network to also identify the subject and object in the convolutional layers instead of solely in the fully-connected layers.
	\end{abstract}
		
	\section{Introduction}
	Understanding the image's content is one of the main goals of computer vision. This may be used\footnote{Information of how the fine-tuned models may be used to address these issues is provided in Appendix \ref{sec:app_usecases}} for \textit{automatic image caption generation}, \textit{image retrieval}, \textit{visual question answering}, and \textit{accessibility}. To achieve this, machines must be capable of understanding the image contents -- i.e. detecting objects (\ac{aka} object detection) that are related to each other via a predicate (\ac{aka} visual relationship detection). Taking advantage of \acp{cnn} and large-scale datasets (such as ImageNet \citep{deng2009imagenet}), object detection has achieved great success but the detection of visual relationships is still in its infancy. This is mainly due to the long-tail distribution (where several predicates have very few instances), interpreting a 3D world from a 2D image, and intra-class diversity -- thus, making it a very difficult task \citep{dai2017detecting, deng2009imagenet, johnson2015image, lu2016visual}.
	\\
	
		\begin{figure}[t]
		\centering
		\mycaptions
		\newcommand{\sampleimageswidth}{0.3}
		\setlength{\tabcolsep}{2pt}
		\begin{tabular}{ccc}
			\subfloat[Union]{\includegraphics[width=\sampleimageswidth\columnwidth]{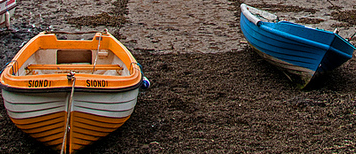}} & 
			\subfloat[Union-WB]{\includegraphics[width=\sampleimageswidth\columnwidth]{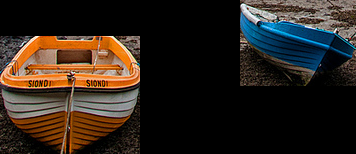}} & 
			\subfloat[Union-WB-SC]{\includegraphics[width=\sampleimageswidth\columnwidth]{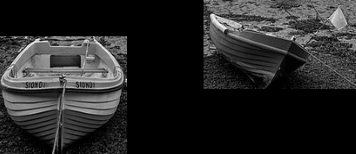}} \\
			
			\subfloat[Union-WB-B]{\includegraphics[width=\sampleimageswidth\columnwidth]{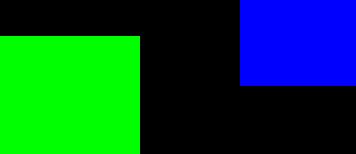}} & 
			\subfloat[Union-WB-B-SC]{\includegraphics[width=\sampleimageswidth\columnwidth]{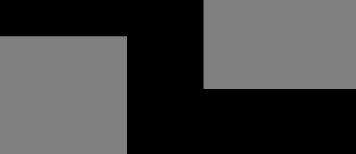}} & \subfloat[Union-WB-and-B]{\includegraphics[width=\sampleimageswidth\columnwidth]{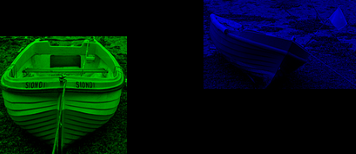}} \\
			
			\subfloat[Segment]{\includegraphics[width=\sampleimageswidth\columnwidth]{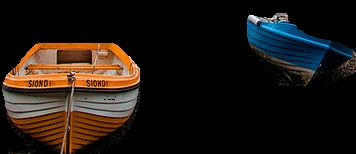}} & \subfloat[Segment-B]{\includegraphics[width=\sampleimageswidth\columnwidth]{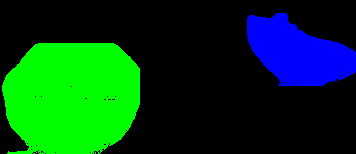}} & 
			\subfloat[Blur-Sigma3]{\includegraphics[width=\sampleimageswidth\columnwidth]{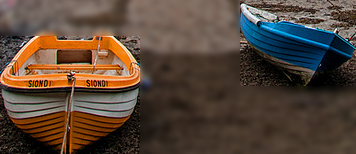}} \\
			
			\subfloat[Blur-Sigma5]{\includegraphics[width=\sampleimageswidth\columnwidth]{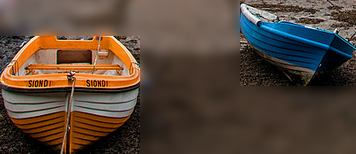}} & \subfloat[Blur-Sigma7]{\includegraphics[width=\sampleimageswidth\columnwidth]{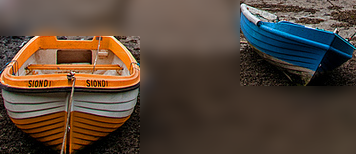}} & \\
			
		\end{tabular}
		\caption{Sample images of the preprocessing methods used.}
		\label{fig:dataset_samples}
	\end{figure}

	\citet{sadeghi2011recognition} addressed the \ac{vrd} problem as a \ac{slc} task where each \textit{triplet} -- composed of the \textit{subject}, linked to an \textit{object} via the \textit{predicate}\footnote{In the Stanford \ac{vrd} dataset, a predicate may be a verb (\ac{eg} wear), spatial (\ac{eg} next to), preposition (\ac{eg} with), comparative (\ac{eg} taller than), action (\ac{eg} kick),	or preposition phrase (\ac{eg} drive on). \citep{lu2016visual}} -- as a unique class. This is referred to as a \textit{visual phrase} (\ac{eg} \enquote{bicycle next to car}). Later, \citet{lu2016visual} pointed out that it is not feasible due to the lack of instances for each visual phrase and the class exponential growth. Therefore, they predicted each one -- the subject, predicate, and object -- separately using three fine-tuned \ac{vgg} models. Furthermore, to filter out irrelevant triplets, language priors obtained from the Stanford \ac{vrd} training dataset were also used. Other research used a novel feature extraction layer \citep{zhang2017visual}; a spatial module to extract spatial configurations and a \ac{drnet} for statistical dependencies \citep{dai2017detecting}; and knowledge distillation using both internal and external linguistic knowledge \citep{yu2017visual}.
	\\
		
	As \citet{zhu2018deep} pointed out, mainly the Union method is used or the spatial configuration is used without any visual information. Additionally, \citet{zhang2016yin} found that in \ac{vqa} -- where vision and language are intertwined -- the answers were not grounded on the image content but based on \textit{superficial performance} (\ac{eg} predicting the answer based on the question only). This can also be applied for the \ac{vrd} scenario, where both vision and language are being used, and models based on statistical dependencies are being employed \citep{dai2017detecting, lu2016visual, zhu2018deep}.
	To ensure that the results are based on the image content, this paper analyses the visual part of the \ac{vrd} problem by focusing solely on predicate prediction\footnote{The ground truth bounding boxes and the respective object categories are provided; the model predicts only the predicate.} -- \ac{ie} the model is only given the image. 
	\citet{zhou2014object} found that the convolutional layers of a \ac{cnn} act as object detectors but previous research mainly fed the Union into the \ac{cnn}. Therefore, not taking full advantage of the convolutional layers as these layers are not capable of determining the location of the respective triplet subject and object. To address these shortcomings, ten preprocessing methods (depicted in Figure  \ref{fig:dataset_samples}) -- Union-WB, Union-WB-SC, Union-WB-B, Union-WB-B-SC, Union-WB-and-B, Segment, Segment-B, Blur-Sigma3, Blur-Sigma5, Blur-Sigma7 -- were used to determine if better results could be achieved when compared to the Union method.
	\\
	
	The main contributions\footnote{To the best of our knowledge they have never been used for the detection of visual relationships.} of this paper are: 
	(1) a novel analysis of the visual module using preprocessing methods;
	(2) the novel usage of segmentations for the detection of visual relationships;
	(3) the novel usage of activation maximisation to study the \ac{vrd} problem; and
	(4) the per-predicate and error analysis.
	\\
	
	\begin{figure*}[t]
		\centering
		\includegraphics[width=\linewidth]{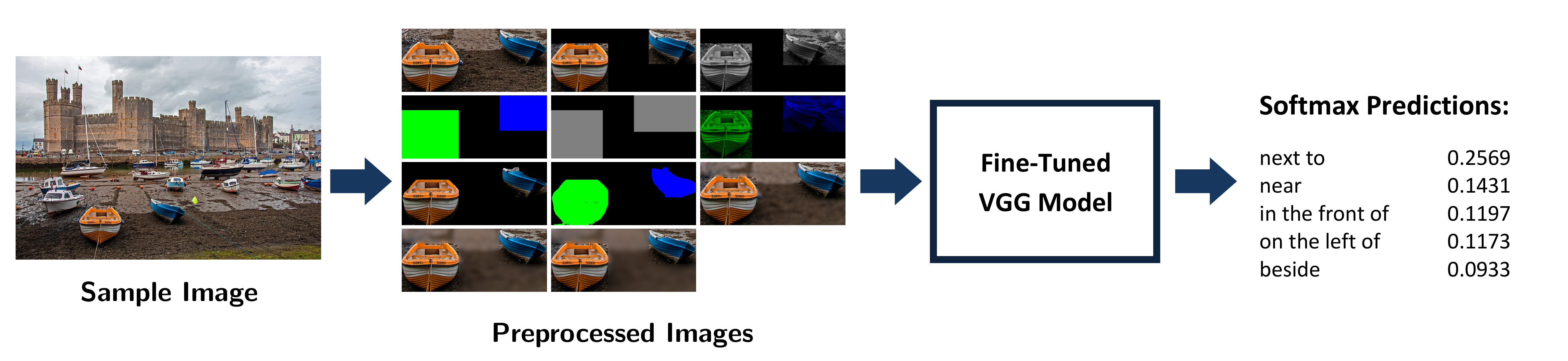}
		\caption{The proposed architecture overview. First, the Stanford \ac{vrd} dataset images and their respective annotations are extracted for each preprocessing method. Each set of preprocessed images (\ac{eg} the Union-WB-B method) is then used to fine-tune both the \ac{vgg16} and the \ac{vgg19} models for predicate prediction. These models were then evaluated using Recall@1, per-predicate recall, activation maximisation, \acp{cam}, and error analysis.}
		\label{fig:model_overview}
	\end{figure*}

	\section{Related Work}
	\subsection{Object Detection}
	Object detection -- the kernel of \ac{vrd} -- is the process of detecting objects pertaining to a specific class  -- \ac{slc} -- and their location \cite{amit2014object, russakovsky2015imagenet}. Recently, \acp{cnn} have been used for object detection to map the image pixels to one or more classes without the need for hand-designed features \citep{sermanet2013overfeat}. \citet{girshick2014rich} developed a \ac{rcnn} which replaces \citet{sermanet2013overfeat} sliding window approach. \ac{fastrcnn} \citep{girshick2015fast} used \ac{spp} \citep{he2014spatial} which eliminates the need for feature caching and enables all parameters to be trainable \citep{girshick2015fast}. Later, \ac{fasterrcnn} \citep{ren2015faster} used a \ac{rpn} \citep{ren2015faster}, while \ac{maskrcnn} \citep{he2017mask} also generated the object segmentations \citep{he2017mask}.
	\\

	\subsection{Visual Relationship Detection}
	The concept of \ac{vrd} is that given an image, the linguistically-correct predicate is predicted. Contrary to object detection, the images within each class are highly uncorrelated and therefore it is a challenge to learn from very few samples \citep{lu2016visual}. To address this problem, two main types of architectures have been developed -- joint and separate. The former considers each triplet as a unique class ($O(N^{2}R)$ where $N$ = \textit{objects}, and $R$ = \textit{predicates}) \citep{sadeghi2011recognition}, while the latter predicts the subject, predicate and objects separately ($O(N+R)$) \citep{lu2016visual, dai2017detecting, yu2017visual, zhu2018deep}.
	\\
	
	\citet{sadeghi2011recognition} predicted each \textit{triplet} as a unique class where the within-class variance is limited and therefore enables learning from few samples. As \citet{lu2016visual} noted, the issue is that: (1) it is difficult to collect enough samples for each \textit{visual phrase} (\ac{eg} "person riding horse"), and (2) the exponential growth of the number of visual phrases when new objects are added to the dataset -- therefore, it is not capable of generalising. To overcome these issues, \citet{lu2016visual} predicted the subject, predicate, and object separately using three fine-tuned \ac{vgg} models initialised using ImageNet \citep{deng2009imagenet} weights. Furthermore, they also used linguistic knowledge from the Stanford \ac{vrd} training dataset to remove improbable predictions and enhance the zero-shot\footnote{Using a separate dataset, evaluate the model to predict unseen triplets} scenario. They achieved this by mapping similar relations (\ac{eg} \textit{\textless person--ride--elephant\textgreater} and \textit{\textless person--ride--horse\textgreater}) close to each in an embedding space -- thus, achieving an 11\% improvement \citep{lu2016visual}. 
	\\
	
	\citet{zhang2017visual} used a novel feature extraction layer (composed of class probabilities, bounding boxes location and scales) to simultaneously predict the objects and predicates using end-to-end training. Additionally, they also allowed knowledge transfer by replacing \ac{fasterrcnn}'s \ac{roi} pooling layer with bilinear interpolation \citep{zhang2017visual}. This resulted in better generalisation due to the class probabilities and improved object detection as a result of knowledge transfer \citep{zhang2017visual}. On the other hand, \citet{dai2017detecting} developed a model based on the spatial configuration and statistical dependencies. The former was composed of two binary masks, one for the subject and another for the object, resized to 32x32 pixels, convolved through three convolutional layers and a fully-connected one to generate a 64-D vector representing the spatial configuration (improving the Recall@50 results by approximately 8.5\% -- much higher than \cites{yu2017visual} spatial features improvement) \citep{dai2017detecting}. On the other hand, a novel \ac{drnet} was used to exploit statistical dependencies. Analogous to \cites{lu2016visual} method, better results were achieved when both visual and statistical modules were used -- especially when the \ac{drnet} was used (although no zero-shot results were provided) \cite{dai2017detecting}. \citet{yu2017visual} used a probabilistic model where instead of increasing the dataset size (due to the long-tail distribution), a larger linguistic knowledge base was used. To achieve this internal knowledge from the Stanford \ac{vrd} and the \ac{vg} datasets and external knowledge from the \textit{Wikipedia} dump\footnote{Used to overcome having triplets with a probability of zero.} were employed. From their research, \citet{yu2017visual} found that better results may be achieved by using a larger linguistic knowledge base. \citet{zhu2018deep} noted that research mainly used the Union method which might contain noise or the object's positional information without any visual information. Therefore, they used \citet{dai2017detecting} spatial module, a feature-level (learning discriminative features) and label-level (capture statistical dependencies) triplet prediction. \citet{zhu2018deep} found that better discriminative features can be attained if the subject and object feature vectors were used. Additionally, analogous to previous methods the statistical dependencies yield better results \citep{zhu2018deep}.
	\\

	\subsection{Understanding \acp{cnn}}
	Due to the large number of interacting/non-linear parts, \acp{cnn} are generally referred to as a \enquote{black box} since they automatically learn useful features from the input to generate the expected output \citep{yosinski2015understanding}. To close this gap Deep Visualisation methods such as \textit{activation maximisation} and \textit{\acp{cam}} have been used. The former numerically -- through back-propagation -- generates an image which depicts the neuron activations for a specific class \citep{nguyen2016multifaceted, simonyan2013deep}. On the other hand, -- since the convolutional layers of a \ac{cnn} act as object detectors -- \acp{cam} are used to visualise the activations (using a single forward pass) at a certain layer when a specific input (an image) is fed into the \ac{cnn} \citep{zhou2016learning, zhou2014object}.
	\\

	\section{Methodology}
	\citet{zhang2016yin} found that in \ac{vqa}, where both vision and language are used, the predictions were not grounded on the image content; while \citet{zhou2016learning} found that the convolutional layers of a \ac{cnn} act as object detectors. Therefore, since research carried out in the field of \ac{vrd} mainly incorporates linguistic knowledge (statistical dependencies and probabilistic models) \citep{lu2016visual, yu2017visual, dai2017detecting, zhu2018deep} or used spatial features as a vector \citep{dai2017detecting, zhu2018deep, zhang2017visual} and mainly feed the Union to generate the predictions, this paper focuses solely on the visual module predicate prediction capability, with the aim of exploiting implicit language derived knowledge. As depicted in Fig. \ref{fig:model_overview}, the experimental setup is composed of a fine-tuned \ac{vgg} model initialised using ImageNet weights. As input it takes a preprocessed image (containing only two annotated objects) and predicts the respective predicate.
	\\
	
	First, the Stanford \ac{vrd} dataset images were preprocessed (Fig. \ref{fig:dataset_samples} depicts sample images) for each method -- the Union, Union-WB, Union-WB-SC, Union-WB-B, Union-WB-B-SC, Union-WB-and-B, Segment, Segment-B, Blur-Sigma3, Blur-Sigma5, and Blur-Sigma7. Since the Stanford \ac{vrd} dataset does not contain any segmentations, the intersection of the Stanford \ac{vrd} and the \ac{mscoco} datasets was used to extract the segmentations using \ac{maskrcnn} \citep{he2017mask}. To ensure that the correct object was segmented, only the respective bounding box content was fed into \ac{maskrcnn} and the predicted class was compared with the ground truth one. Additionally, to ensure a fair setting, the bounding boxes provided by \ac{maskrcnn} were used instead of those provided by the Stanford \ac{vrd} dataset.
	\\
	
	The following is an explanation of each preprocessing method used for predicate predication:
	\begin{itemize}
		\item \textbf{Union} --- the smallest bounding box that encloses both the subject and the object bounding boxes;
		\item \textbf{Union-WB} --- the background (\ac{ie} the pixels not within the subject or object bounding boxes) of the respective Union image is removed;
		\item \textbf{Union-WB-SC} --- the single channel version (\ac{ie} grayscale image) of the Union-WB method;
		\item \textbf{Union-WB-B}\footnote{Inspired by \cites{dai2017detecting} spatial mask} --- the object binary mask was set on the first channel, while the subject binary mask was set on the second channel;
		\item \textbf{Union-WB-B-SC} --- the Union-WB-B method on a single channel (\ac{ie} grayscale image);
		\item \textbf{Union-WB-and-B} --- the Union-WB method but the single channel bounding boxes were used instead of the binary masks;
		\item \textbf{Segment} --- the Union was extracted and the pixels not within the subject or object segmentations were zeroed-out;
		\item \textbf{Segment-B} --- analogous to the Union-WB-B method but the respective segmentations were used instead of the bounding boxes;
		\item \textbf{Blurring} ---	\citet{gu2015recent} found that the lower layers of a \ac{cnn} act as edge detectors. Therefore, instead of removing the background (\ac{eg} Union-WB), three blurring variations were used to reduce attention in this area\footnote{Blurring the background pixels}. To achieve this, a Gaussian low-pass filter using standard deviations\footnote{For both the $x$ and $y$ directions and chosen through manual testing and as employed by \citet{gonzalez2017digital}} 3 (\textbf{Blur-Sigma3}), 5 (\textbf{Blur-Sigma5}), and 7 (\textbf{Blur-Sigma7})  was used to blur the background using the kernels $19\times19$, $31\times31$, and $43\times43$ respectively.
	\end{itemize}

	\citet{girshick2014rich} found that using a large dataset (such as ImageNet) to pre-train a \ac{cnn} and then fine-tune it using the desired dataset was very effective. Consequently, a pre-trained \ac{vgg16}/\ac{vgg19} model, initialised using ImageNet weights \citep{deng2009imagenet} was fine-tuned using mini-batch \ac{sgd} for predicate prediction. This was achieved by first normalising the image pixels between zero and one, without any image augmentations\footnote{\ac{eg} when flipped horizontally, \textit{on the left of} becomes \textit{on the right of} and vice versa}. Subsequently, the images were resized to $224 \times 224$ pixels, the respective \ac{vgg} model was then initialised and all layers excluding the fully-connected layers were set to non-trainable -- the weights were not fine-tuned. Using \textit{categorical cross-entropy}, a batch size of 10, Nesterov momentum set to 0.9, and a learning rate of 0.001 the model was fine-tuned for five epochs. Once complete, the last convolutional block layers and the fully-connected layers were set to trainable and the model was fine-tuned for five epochs. Finally, using a learning rate of 0.00001, the model was fine-tuned for five more epochs. The fine-tuned model was then evaluated as explained in the following section.
	\\

	\section{Experiments}
	As already stated, the fine-tuned models were evaluated for predicate prediction -- \ac{ie} the model was provided the ground truth subject and object bounding boxes and categories (\ac{eg} horse) -- on the Stanford \ac{vrd} dataset. Therefore, the model only had to predict the correct predicate -- no object detection was used \citep{dai2017detecting, lu2016visual, yu2017visual, zhang2017visual}.

	\begin{table}
	\begin{center}
		\caption{Results on the Stanford \ac{vrd} test set -- Tukey's HSD post-hoc groupings  (R@1 means Recall@1)}
		\label{tab:res_fulltest_ttest}
		\begin{tabular}{c|c|c|c}
			Architecture & Method & Mean R@1 & Group \\
			\hline
			\ac{vgg16} & Union-WB-and-B & 50.26 & A \\
			\ac{vgg19} & Union-WB-and-B & 50.16 & A \\
			\ac{vgg16} & Union-WB-B & 48.03 & B \\
			\ac{vgg19} & Union-WB-B & 47.93 & B \\
			\ac{vgg16} & Union-WB & 39.37 & C \\
			\ac{vgg16} & Union-WB-SC & 39.34 & C \\
			\ac{vgg19} & Union-WB-SC & 39.23 & C \\
			\ac{vgg19} & Union-WB & 39.04 & C \\
			\ac{vgg16} & Blur-Sigma7 & 37.74 & D \\
			\ac{vgg19} & Blur-Sigma7 & 37.58 & D \\
			\ac{vgg19} & Blur-Sigma5 & 37.46 & D \\
			\ac{vgg16} & Blur-Sigma5 & 37.27 & D,E \\
			\ac{vgg19} & Blur-Sigma3 & 36.84 & E,F \\
			\ac{vgg16} & Blur-Sigma3 & 36.75 & F \\
			\ac{vgg16} & Union & 35.52 & G \\
			\ac{vgg19} & Union & 35.13 & G \\
			\ac{vgg16} & Union-WB-B-SC & 32.54 & H \\
			\ac{vgg19} & Union-WB-B-SC & 32.40 & H \\
			\end{tabular}
		\end{center}
	\end{table}

	\begin{table}
		\begin{center}
			\caption{Results on the Stanford \ac{vrd} zero-shot set -- Tukey's HSD post-hoc groupings  (R@1 means Recall@1)}
			\label{tab:res_fullzero_ttest}
			\begin{tabular}{c|c|c|c}
				Architecture & Method & Mean R@1 & Group \\
				\hline
				\ac{vgg19} &	 Union-WB-and-B	&	23.80	&	A \\
				\ac{vgg16} &	 Union-WB-and-B	&	23.75	&	A \\
				\ac{vgg16} &	Union-WB-B	&	21.98	&	B \\
				\ac{vgg19} &	Union-WB-B	&	21.85	&	B \\
				\ac{vgg16} &	Union-WB	&	17.40	&	C \\
				\ac{vgg19} &	Union-WB	&	17.31	&	C \\
				\ac{vgg19} &	Union-WB-SC	&	17.02	&	C \\
				\ac{vgg16} &	Union-WB-SC	&	16.80	&	C,D \\
				\ac{vgg19} &	Blur-Sigma7	&	15.76	&	D,E \\
				\ac{vgg19} &	Blur-Sigma5	&	15.47	&	E \\
				\ac{vgg16} &	Blur-Sigma7	&	15.38	&	E \\
				\ac{vgg16} &	Blur-Sigma5	&	15.00	&	E \\
				\ac{vgg19} &	Blur-Sigma3	&	14.94 &	E,F \\
				\ac{vgg16} &	Blur-Sigma3	&	13.81	&	F,G \\
				\ac{vgg16} &	Union-WB-B-SC	&	13.55	&	G \\
				\ac{vgg19} &	Union-WB-B-SC	&	13.55	&	G \\
				\ac{vgg19} &	Union	&	11.29	&	H \\
				\ac{vgg16} &	Union	&	10.64	&	H \\
			\end{tabular}
		\end{center}
	\end{table}
	
	\begin{table}
		\begin{center}
			\caption{Results on the Stanford \ac{vrd} test subset -- Tukey's HSD post-hoc groupings  (R@1 means Recall@1)
				\label{tab:res_test_ttest}}
			\begin{tabular}{c|c|c|c}
				\textbf{Architecture} & \textbf{Method} & \textbf{Mean R@1} & \textbf{Group} \\
				\hline
				\ac{vgg19} & Segment-B & 29.11 & A \\
				\ac{vgg16} & Union-WB-B & 28.99 & A \\
				\ac{vgg19} & Union-WB-B & 28.80 & A \\
				\ac{vgg16} & Segment-B & 28.36 & A \\
				\ac{vgg16} & Union-WB & 26.32 & B \\
				\ac{vgg19} & Blur-Sigma7 & 25.62 & B, C \\
				\ac{vgg19} & Union-WB & 25.60 & B, C \\
				\ac{vgg16} & Blur-Sigma7 & 25.42 & B, C \\
				\ac{vgg16} & Blur-Sigma3 & 25.41 & B, C \\
				\ac{vgg19} & Blur-Sigma5 & 25.37 & B, C \\
				\ac{vgg16} & Blur-Sigma5 & 25.30 & B, C \\
				\ac{vgg19} & Blur-Sigma3 & 25.27 & B, C \\
				\ac{vgg19} & Segment & 25.11 & B, C \\
				\ac{vgg16} & Segment & 25.09 & B, C \\
				\ac{vgg19} & Union & 24.62 & C \\
				\ac{vgg16} & Union & 24.52 & C \\
			\end{tabular}
		\end{center}
	\end{table}
	
	\begin{table}
		\begin{center}
			\caption{Results on the Stanford \ac{vrd} zero-shot subset -- Tukey's HSD post-hoc groupings  (R@1 means Recall@1)
				\label{tab:res_zero_ttest}}
			\begin{tabular}{c|c|c|c}
				\textbf{Architecture} & \textbf{Method} & \textbf{Mean R@1} & \textbf{Group} \\
				\hline
				\ac{vgg16} & Union-WB-B & 14.54 & A \\
				\ac{vgg19} & Union-WB-B & 13.88 & A \\
				\ac{vgg16} & Segment-B & 12.50 & B \\
				\ac{vgg19} & Segment-B & 12.14 & B \\
				\ac{vgg19} & Union-WB & 10.05 & C \\
				\ac{vgg19} & Segment & 10.05 & C \\
				\ac{vgg19} & Union & 10.02 & C \\
				\ac{vgg16} & Segment & 9.95 & C \\
				\ac{vgg19} & Blur-Sigma3 & 9.82 & C \\
				\ac{vgg16} & Union-WB & 9.81 & C \\
				\ac{vgg19} & Blur-Sigma7 & 9.64 & C \\
				\ac{vgg16} & Union & 9.62 & C \\
				\ac{vgg16} & Blur-Sigma5 & 9.60 & C \\
				\ac{vgg19} & Blur-Sigma5 & 9.48 & C \\
				\ac{vgg16} & Blur-Sigma7 & 9.42 & C \\
				\ac{vgg16} & Blur-Sigma3 & 9.36 & C \\
			\end{tabular}
		\end{center}
	\end{table}

	\begin{table*}
		\begin{center}
			\caption[\ac{vgg13} testing dataset per-predicate results]{The per-predicate Recall@1 results for the testing dataset when the \ac{vgg16} architecture was used. Only the predicates that obtained a Recall@1 result greater than 0 are listed. The highest result for each method is italicised, while the highest result for each predicate is marked in bold.
				\label{tab:res_full16test_pred}}
			\begin{tabular}{c|c|c|c|c|c|c|c|c|c}
				\rot{60}{Predicate} & \rot{60}{Union} & \rot{60}{Union-WB} & \rot{60}{Union-WB-SC} & \rot{60}{Union-WB-B} & \rot{60}{Union-WB-B-SC} & \rot{60}{Union-WB-and-B} & \rot{60}{Blur-Sigma3} & \rot{60}{Blur-Sigma5} & \rot{60}{Blur-Sigma7} \\
				\hline
				above 	&	60.21	&	67.32	&	60.77	&	70.24	&	68.31	&	\textbf{70.49}	&	58.20	&	59.87	&	62.40	\\
				behind 	&	27.15	&	33.05	&	14.98	&	19.00	&	33.93	&	\textbf{34.77}	&	28.12	&	24.81	&	27.91	\\
				below 	&	0.00	&	0.00	&	0.00	&	\textbf{0.17}	&	0.00	&	\textbf{0.17}	&	0.00	&	0.00	&	0.00	\\
				in 	&	0.00	&	\textbf{3.45}	&	0.00	&	0.00	&	1.79	&	0.51	&	2.90	&	1.24	&	0.83	\\
				in the front of 	&	0.25	&	6.38	&	0.56	&	12.31	&	5.87	&	\textbf{28.00}	&	1.94	&	5.00	&	2.75	\\
				near 	&	0.00	&	1.53	&	0.68	&	1.02	&	0.17	&	0.34	&	0.85	&	1.70	&	\textbf{2.13}	\\
				next to 	&	35.29	&	47.93	&	\textbf{57.94}	&	\textbf{57.94}	&	46.99	&	54.54	&	39.51	&	40.49	&	42.24	\\
				on 	&	61.66	&	67.15	&	40.22	&	\textbf{87.15}	&	67.81	&	85.24	&	64.56	&	65.95	&	65.61	\\
				on the left of	&	0.00	&	0.00	&	0.00	&	\textbf{1.36}	&	0.00	&	0.00	&	0.00	&	0.00	&	0.00	\\
				on the right of	&	0.00	&	0.00	&	0.00	&	\textbf{0.89}	&	0.00	&	0.00	&	0.00	&	0.00	&	0.00	\\
				under 	&	0.60	&	0.06	&	0.00	&	58.05	&	0.00	&	\textbf{61.52}	&	0.06	&	0.06	&	0.00	\\
				has 	&	0.25	&	2.29	&	0.00	&	0.04	&	2.43	&	\textbf{21.66}	&	1.16	&	1.20	&	1.83	\\
				hold 	&	0.00	&	0.52	&	0.00	&	1.72	&	0.17	&	\textbf{3.97}	&	0.00	&	0.00	&	0.69	\\
				wear 	&	 \textit{85.1} 	&	 \textit{85.89} 	&	 \textit{86.90} 	&	 \textbf{\textit{96.06}}	&	 \textit{84.87} 	&	 \textit{88.38} 	&	 \textit{85.8} 	&	 \textit{86.62} 	&	 \textit{86.79}	\\
			\end{tabular}
		\end{center}
	\end{table*}

	\begin{figure}
		\centering
		\mycaptions
		\newcommand{\contwidth}{0.25}
		\subfloat[Above]{\includegraphics[width=\contwidth\linewidth]{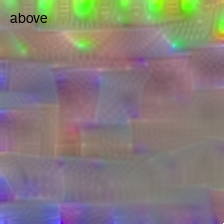}}
		\qquad
		\subfloat[Above -- Subject Channel]{\includegraphics[width=\contwidth\linewidth]{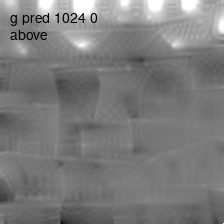}}
		\qquad
		\subfloat[Above -- Object Channel]{\includegraphics[width=\contwidth\linewidth]{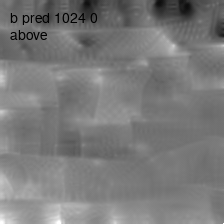}}
		\\
		\subfloat[Below]{\includegraphics[width=\contwidth\linewidth]{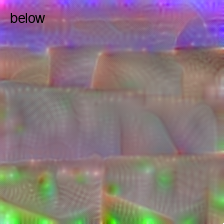}}
		\qquad
		\subfloat[Below -- Subject Channel]{\includegraphics[width=\contwidth\linewidth]{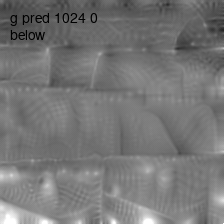}}
		\qquad
		\subfloat[Below -- Object Channel]{\includegraphics[width=\contwidth\linewidth]{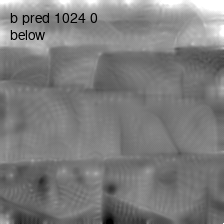}}
		\\
		\subfloat[Beneath]{\includegraphics[width=\contwidth\linewidth]{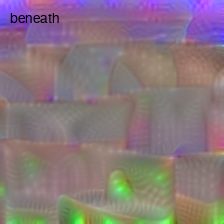}}
		\qquad
		\subfloat[Beneath -- Subject Channel]{\includegraphics[width=\contwidth\linewidth]{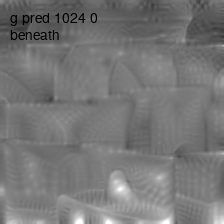}}
		\qquad
		\subfloat[Beneath -- Object Channel]{\includegraphics[width=\contwidth\linewidth]{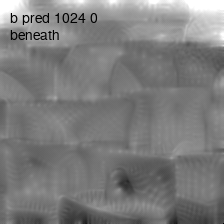}}
		\\
		\subfloat[Under]{\includegraphics[width=\contwidth\linewidth]{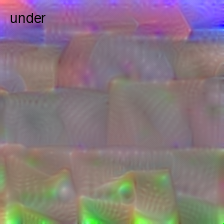}}
		\qquad
		\subfloat[Under -- Subject Channel]{\includegraphics[width=\contwidth\linewidth]{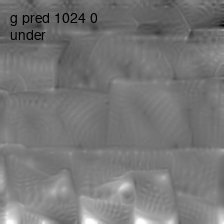}}
		\qquad
		\subfloat[Under -- Object Channel]{\includegraphics[width=\contwidth\linewidth]{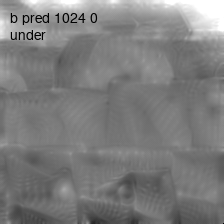}}
		
		\caption[Union-WB-B method activation maximisation results]{The Union-WB-B method activation maximisation results. These images show that the model learnt what \textit{above}, \textit{below}, \textit{beneath}, and \textit{under} mean when the Union-WB-B method was used. Images \textit{a}, \textit{d}, \textit{g}, and \textit{j} are a \ac{rgb} image, while images \textit{b-c}, \textit{e-f}, \textit{h-i}, and \textit{k-l} are the \textit{one-channel} analysis results -- \ac{ie}, images \textit{b}, \textit{e}, \textit{h}, and \textit{k} are the subject channel, while \textit{c}, \textit{f}, \textit{i}, and \textit{l} are the object channel. (The same channels used when creating the Union-WB-B dataset)
			\label{fig:res_actmax_abovebelow_bb}}
	\end{figure}
		
		\begin{figure}[h]
		\centering
		\mycaptions
		\newcommand{\contwidth}{0.25}
		\subfloat[On the left of]{\includegraphics[width=\contwidth\linewidth]{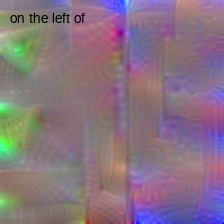}}
		\qquad
		\subfloat[On the left of -- Subject Channel]{\includegraphics[width=\contwidth\linewidth]{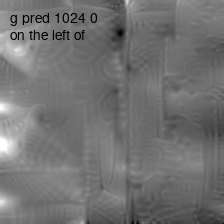}}
		\qquad
		\subfloat[On the left of -- Object Channel]{\includegraphics[width=\contwidth\linewidth]{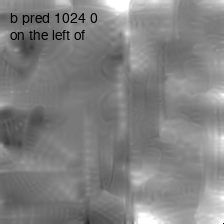}}
		\\
		\subfloat[On the right of]{\includegraphics[width=\contwidth\linewidth]{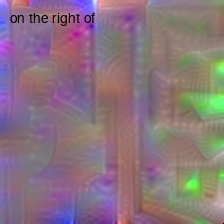}}
		\qquad
		\subfloat[On the right of -- Subject Channel]{\includegraphics[width=\contwidth\linewidth]{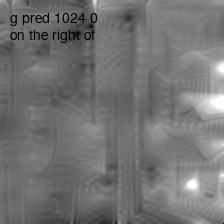}}
		\qquad
		\subfloat[On the right of -- Object Channel]{\includegraphics[width=\contwidth\linewidth]{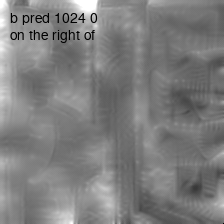}}
		
		\caption[Union-WB-B method activation maximisation results]{The Union-WB-B method activation maximisation results. As in Figure \ref{fig:res_actmax_abovebelow_bb}, these images show that the model learnt what \textit{on the left of} and \textit{on the right of} mean, when the Union-WB-B method was used. Images \textit{a} and \textit{d} are a \ac{rgb} image, while images \textit{b}, \textit{c}, \textit{e} and \textit{f} are the \textit{one-channel} analysis results -- \ac{ie} images \textit{b} and \textit{e} are the subject channel, while \textit{c} and \textit{f} are the object channel. (The same channels used when creating the Union-WB-B dataset)
			\label{fig:res_actmax_leftright_wb}}
	\end{figure}
	
	\begin{figure}
		\centering
		\mycaptions
		\newcommand{\contwidth}{0.25}
		\subfloat[Drive]{\includegraphics[width=\contwidth\linewidth]{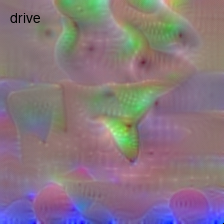}}
		\qquad
		\subfloat[On]{\includegraphics[width=\contwidth\linewidth]{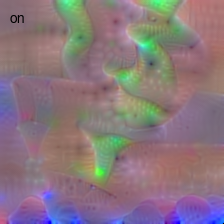}}
		\\
		\subfloat[On the top of]{\includegraphics[width=\contwidth\linewidth]{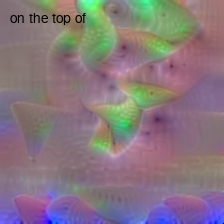}}
		\qquad
		\subfloat[Ride]{\includegraphics[width=\contwidth\linewidth]{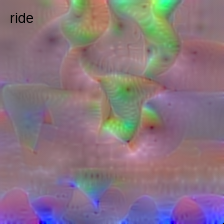}}
		\qquad
		\subfloat[Sit On]{\includegraphics[width=\contwidth\linewidth]{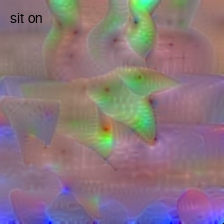}}

		\caption[Segment-B method activation maximisation results]{The Segment-B method activation maximisation results. These activation maximisation results show that the model has learnt the same features.
			\label{fig:res_actmax_segb}}
	\end{figure}

	\begin{figure}
		\centering
		\mycaptions
		\subfloat[Union-WB \ac{vgg16}]{\includegraphics[width=0.4\linewidth]{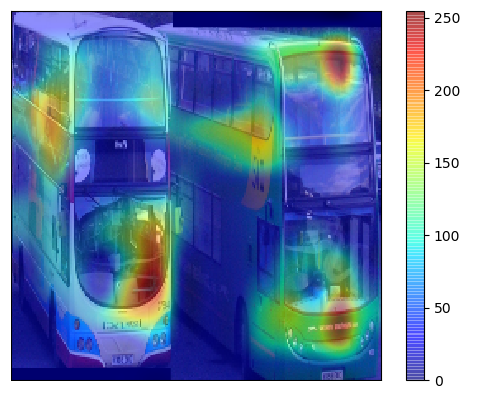}}
		\quad
		\subfloat[Union-WB \ac{vgg19}]{\includegraphics[width=0.4\linewidth]{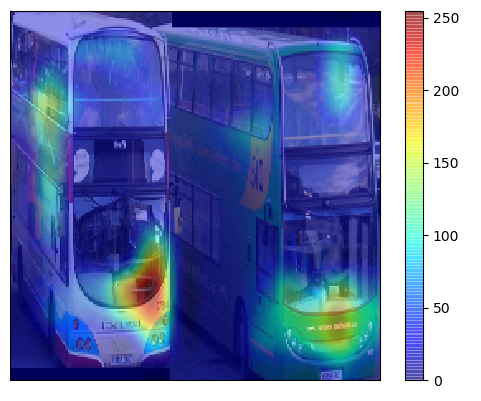}}

		\caption{Triplet \textit{\textless bus--next to--bus\textgreater} \ac{cam} example, where the \ac{vgg16} model predicted the predicate \textit{near}, while the \ac{vgg19} model predicted the predicate \textit{next to}; furthermore the \ac{vgg16} paid more attention to the distance between the two buses than the \ac{vgg19}.}
		\label{fig:res_cam_unionwb}
	\end{figure}
	
	\begin{figure}
		\centering
		\mycaptions
		\subfloat[Blur-Sigma3 \ac{vgg16}]{\includegraphics[width=0.4\linewidth]{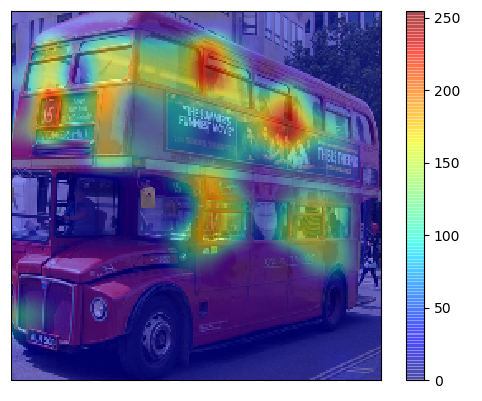}}
		\quad
		\subfloat[Blur-Sigma3 \ac{vgg19}]{\includegraphics[width=0.4\linewidth]{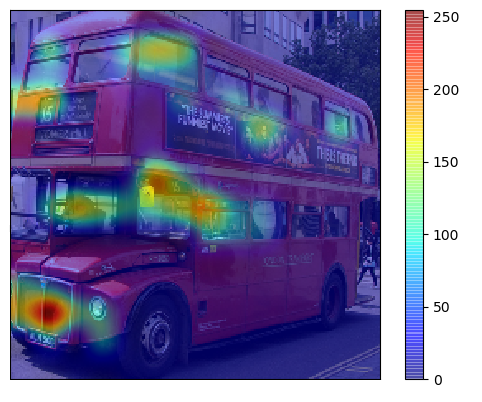}}

		\caption{Triplet \textit{\textless person--drive--bus\textgreater} \ac{cam} example, where both the \ac{vgg16} and \ac{vgg19} models predicted the predicate \textit{in the front of}; while as depicted in figure b, the \ac{vgg19} paid attention to the driver.}
		\label{fig:res_cam_bs3}
	\end{figure}
	
	\begin{table}
		\begin{center}
			\caption{The error analysis findings}
			\label{tab:res_erroranalysis}
			\begin{tabular}{c|c}
				Type & Percentage \\
				\hline
				Alternative Predicates & 44\% \\
				Different \acp{pov} & 16\% \\
				Incorrect Prediction & 11\% \\
				Linguistic Error & 9\% \\
				Phrases & 8\% \\
				Synonyms & 6\% \\
				Incorrect Annotation & 6\% \\
				Background Objects & 5\% \\
			\end{tabular}
		\end{center}
	\end{table}

	\subsection{Evaluation Metrics}
	Once trained, the models were evaluated on two datasets -- the testing and zero-shot datasets -- where, contrary to the testing dataset, the zero-shot dataset contains triplets that were never seen by the model. To evaluate these models Recall@1 (which compares the ground truth predicate with the predicted predicate having the maximum probability \citep{lu2016visual, dai2017detecting}), per-predicate recall, activation maximisation, \acp{cam}, and error analysis were used. Since the dataset annotations are incomplete, precision and \ac{map} were not used \citep{lu2016visual, dai2017detecting}. Moreover, each model was fine-tuned multiple times and then Tukey's HSD post-hoc test (using a p-value of 0.05) -- which compares the groups' mean value while reducing Type I errors \cite{black2009business} -- was used to check for statistical significance \citep{crichton1999information, urdan2011statistics}.
	
	\subsection{Dataset}
	Two versions of the Stanford \ac{vrd} dataset which is "the most widely benchmarked dataset for the relationship detection in real-world images" \cite{krishna2018referring} were used. The first one is the full dataset (composed of 5,000 images, 100 object categories, and 70 predicate categories); while the second one, referred to as the \textit{Stanford \ac{vrd} dataset subset}, is a custom dataset -- composed of the intersection of the Stanford \ac{vrd} and the \ac{mscoco} datasets\footnote{Resulting in 1,943 images, 38 object categories and 64 predicates categories.} -- used to compensate for the lack of segmentations in the Stanford \ac{vrd} dataset.
	
	\subsection{Implementation Details}
	Python and Keras \ac{api} using a Tensorflow backend were used to implement and evaluate the \ac{cnn} models, while the Python \textit{Keras-vis} package was used to generate the activation maximisation images (using 1,024 backpropagation iterations) and the \acp{cam} visualisations. 
	
	\subsection{Results and Discussion}
	\citet{lu2016visual} quote the predicate detection results obtained when using the vision module only (7.11\%). However, this result is not applicable to our case since we train our model end-to-end. On the other hand, \citet{yu2017visual} made use of a softmax output trained end-to-end and obtained a Recall@1 of 34.82\% when the input was the Union. The mean Recall@1 for the Union method\footnote{Presented in Table \ref{tab:res_fulltest_ttest}} (35.52\%, \ac{vgg16}) is very close to the result in \citet{yu2017visual} -- (34.82\%, k=1) -- where a similar architecture was used. 
	\\
	
	\begin{table*}
		\begin{center}
			\caption{Various mean Recall@N results for the top-performing method -- \ac{ie} the Union-WB-and-B method -- and the widely-used Union method. (R@n means Recall@n)}
			\label{tab:res_multiple_recall}
			\begin{tabular}{c|c|c|c|c|c|c|c}
				Method & Dataset & R@1 & R@2 & R@3 & R@5 & R@8 & R@10 \\
				\hline
				Union-WB-and-B & Testing & 50.16 & 66.89 & 75.25 & 83.98 & 90.71 & 93.26 \\
				Union-WB-and-B & Zero-Shot & 23.80 & 36.54 & 47.73 & 63.63 & 77.59 & 82.09 \\
				Union & Testing & 35.13 & 49.99 & 60.43 & 72.18 & 83.23 & 87.54 \\
				Union & Zero-Shot & 11.29 & 20.99 & 29.14 & 42.46 & 58.31 & 68.11\\			
			\end{tabular}
		\end{center}
	\end{table*}

	The Recall@1 results for all methods (excluding the segmentation based methods), trained and tested on the full Stanford \ac{vrd} dataset are given in  Tables \ref{tab:res_fulltest_ttest} and \ref{tab:res_fullzero_ttest} for both the testing and zero-shot datasets.  The results including the segmentations methods for the Stanford \ac{vrd} dataset subset are given in Tables \ref{tab:res_test_ttest} and \ref{tab:res_zero_ttest}, while Table \ref{tab:res_multiple_recall} gives Recall@k, (k=1, 2, ...,10)  results  for the Union-WB-and-B and the Union methods.	The following are some conclusions that can be drawn from these tables: (a) As expected, the full dataset Recall@1 results are higher than those of the subset as a result of more training data. (b) Results obtained using the \ac{vgg16} and \ac{vgg19} models are not significantly different. This suggests that less complex architectures may yield similar accuracies; (c) As the Union bounding box background is blurred, and eventually blacked out, the accuracies increase. This follows the intuition that extra objects can distract the attention on the two objects and confirms similar observations in \citet{zhu2018deep}; (d) There is a significant difference in between the binary (spatial) methods and methods that depict the objects.  For example, there is an increase of 8.66\% and 4.58\% in recall from the Union-WB to the Union-WB-B for the (\ac{vgg16}) testing and zero-shot datasets respectively. (e) Additionally, when both were merged -- \ac{ie} the Union-WB-and-B method, -- even better results were achieved. This is mainly due to the use of different channels for the subject and object, while maintaining the respective subject/object visual features. Therefore, detecting object features (\ac{ie} language features implicit in the image) help in modelling the dependencies. (f) The same pattern can be noted for the zero-shot dataset, indicating that object pairs not found in the training set (\ac{eg} \textit{person} and \textit{elephant}) are probably closer to pairs found in the training dataset (\ac{eg} \textit{person} and \textit{horse}). (g) Referring to Table \ref{tab:res_test_ttest} the Segment-B and Union-WB-B are not statistically  different, while Table \ref{tab:res_zero_ttest} shows that they are statistically different. This could be due to \ac{maskrcnn} splitting the respective objects into separate segments when occluded by other objects and using rectangular areas instead of segmentations.
	\\
			
	Table \ref{tab:res_full16test_pred} tabulates the per-predicate Recall@1 accuracies for all methods tested on the full Stanford \ac{vrd} dataset. It can be noted that the models were mainly detecting \textit{spatial prepositions} followed \textit{verbs}, while there are no \textit{actions} or \textit{comparatives}. The highest recall rates are  shared between the Union-WB-B and Union-WB-and-B methods, with the exceptions of \textit{in} and \textit{near}, where the Union-WB and Blur-Sigma7 methods respectively achieved the highest recall. Additionally, the Union-WB-and-B method predicted the most diverse set at top rank. One would expect to have higher accuracies from the Union-WB-and-B method for the more functionally biased predicates, but this is not always the case; \textit{has} and \textit{hold} are two examples of functionally biased predicates that the Union-WB-and-B achieved better results as it includes both spatial and implicit language knowledge. However, \textit{below}, which is known to be a geometrically biased predicate does not do well. This may be due to being predicated 77.98\% of the time as \textit{under} (a functionally biased predicate). Overall, the predicates \textit{on the left of}, \textit{on the right of} and \textit{near} did not attain a high recall result but were predicted as \textit{next to} -- a topological predicate hence easier to comprehend. On the other hand, \textit{above} which is a projective predicate (and requires more attention) achieved better results. The Union-WB-and-B does significantly better with \textit{behind} and \textit{in front of}, which could be that the implicit language knowledge is being used as a proxy to depth in a 3D world. 
	\\
	
	Error analysis was carried out on the outputs to better understand the sources of error.  Table \ref{tab:res_erroranalysis} summarizes the analysis results of a sample of one-hundred random predictions.  The main source of error (44\%) is  \textit{alternative however plausible predicates}, followed by \textit{different \acp{pov}} (16\%). The former may be addressed by using the sigmoid loss function and binary cross-entropy enabling \ac{mlc}, while the latter may be addressed by using a consistent \ac{pov} -- \ac{ie} from the viewer or the respective object \ac{pov}.
	\\
	
	The activation maximisation results for \textit{above}, \textit{below}, \textit{beneath} and \textit{under} are given in Fig. \ref{fig:res_actmax_abovebelow_bb}.  The subject and object channel images depict the respective activations for the respective channel -- \ac{ie} the green (subject) and blue (object) channels. Through the use of different channels, we note that the model identified the relationship direction correctly. Taking a closer look, \textit{below} has a wider range of acceptance, whilst in \textit{beneath} the subject is closer to the centre and \textit{under} is somewhere in between.
	The direction can also be observed for the predicates \textit{on the left of} and \textit{on the right of} activation maximisation results presented in Fig. \ref{fig:res_actmax_leftright_wb}. Additionally, referring to Fig. \ref{fig:res_actmax_segb}, it can be noted that the \ac{cnn} learnt similar features for predicates -- \textit{drive}, \textit{on}, \textit{on the top of}, \textit{ride}, and \textit{sit on} -- where the subject is located on the object, or in general in the upper half of the union box. This shows that the \ac{cnn} was not capable of identifying the difference between these predicates.
	\\
	
	When reviewing the \acp{cam} results, they are inconclusive since there were no key differences between the methods and models apart from those presented in Fig. \ref{fig:res_cam_unionwb} and \ref{fig:res_cam_bs3}. The former shows that for the Union-WB method, the \ac{vgg16} model paid attention to the distance between the buses, while the \ac{vgg19} did not. Conversely, the latter shows that for the Blur-Sigma3 method the \ac{cnn} paid attention to the driver when the \ac{vgg19} model was used but not for the \ac{vgg16} model.
	\\

	\section{Conclusion}
	The aim of this research was to analyse the visual part of the \ac{vrd} problem by using various image preprocessing methods to enable detection in the convolutional layers of the \ac{cnn}. To ensure that the results were based on the image content, no object detection and linguistic knowledge\footnote{In \ac{vqa} it was found that the answers were not grounded based on the image content.} were used. To achieve this, the pre-trained \ac{vgg16} and \ac{vgg19} models were fine-tuned for predicate prediction. Overall, this resulted in the Union-WB-and-B method achieving the best results due to the use of different channels and maintaining the respective bounding box visual features. Furthermore, activation maximisation, \acp{cam}, and error-analysis were used to determine what the \ac{cnn} has learnt.
	
	\section{Future Work}
	Analogous to how \ac{maskrcnn} generated segmentations, we will train an object detector such as \ac{fasterrcnn} or \ac{maskrcnn} to additionally provide the respective object pose/posture. This is useful for both object detection and visual relationship detection. Furthermore, addressing the 44\% alternative predicates\footnote{Error analysis results presented in Table \ref{tab:res_erroranalysis}}, we will train the models using the sigmoid loss function and binary cross-entropy -- \ac{mlc} -- instead of using the softmax loss function -- \ac{slc} -- thus, predicting multiple valid predicates for each pair of objects. Moreover, we will also analyse how smaller \acp{cnn} affect predicate predication.

	\bibliography{references}

\begin{thebibliography}{28}
\providecommand{\natexlab}[1]{#1}
\providecommand{\url}[1]{\texttt{#1}}
\expandafter\ifx\csname urlstyle\endcsname\relax
  \providecommand{\doi}[1]{doi: #1}\else
  \providecommand{\doi}{doi: \begingroup \urlstyle{rm}\Url}\fi

\bibitem[Amit and Felzenszwalb(2014)]{amit2014object}
Yali Amit and Pedro Felzenszwalb.
\newblock Object detection.
\newblock \emph{Computer Vision: A Reference Guide}, pages 537--542, 2014.

\bibitem[Black(2009)]{black2009business}
K.~Black.
\newblock \emph{Business Statistics: Contemporary Decision Making}.
\newblock Wiley Plus Products Series. John Wiley \& Sons, 2009.
\newblock ISBN 9780470409015.

\bibitem[Crichton(1999)]{crichton1999information}
Nicola Crichton.
\newblock Information point: Tukey multiple comparison test.
\newblock \emph{Journal of Clinical Nursing}, 8:\penalty0 299--304, 1999.

\bibitem[Dai et~al.(2017)Dai, Zhang, and Lin]{dai2017detecting}
Bo~Dai, Yuqi Zhang, and Dahua Lin.
\newblock Detecting visual relationships with deep relational networks.
\newblock In \emph{Computer Vision and Pattern Recognition (CVPR), 2017 IEEE
  Conference on}, pages 3298--3308. IEEE, 2017.

\bibitem[Deng et~al.(2009)Deng, Dong, Socher, Li, Li, and
  Fei-Fei]{deng2009imagenet}
Jia Deng, Wei Dong, Richard Socher, Li-Jia Li, Kai Li, and Li~Fei-Fei.
\newblock Imagenet: A large-scale hierarchical image database.
\newblock In \emph{Computer Vision and Pattern Recognition, 2009. CVPR 2009.
  IEEE Conference on}, pages 248--255. IEEE, 2009.

\bibitem[Girshick(2015)]{girshick2015fast}
Ross Girshick.
\newblock Fast r-cnn.
\newblock In \emph{Proceedings of the IEEE international conference on computer
  vision}, pages 1440--1448, 2015.

\bibitem[Girshick et~al.(2014)Girshick, Donahue, Darrell, and
  Malik]{girshick2014rich}
Ross Girshick, Jeff Donahue, Trevor Darrell, and Jitendra Malik.
\newblock Rich feature hierarchies for accurate object detection and semantic
  segmentation.
\newblock In \emph{Proceedings of the IEEE conference on computer vision and
  pattern recognition}, pages 580--587, 2014.

\bibitem[Gonzalez and Woods(2017)]{gonzalez2017digital}
R.C. Gonzalez and R.E. Woods.
\newblock \emph{Digital Image Processing}.
\newblock Pearson, 2017.
\newblock ISBN 9780133356724.

\bibitem[Gu et~al.(2015)Gu, Wang, Kuen, Ma, Shahroudy, Shuai, Liu, Wang, Wang,
  Wang, et~al.]{gu2015recent}
Jiuxiang Gu, Zhenhua Wang, Jason Kuen, Lianyang Ma, Amir Shahroudy, Bing Shuai,
  Ting Liu, Xingxing Wang, Li~Wang, Gang Wang, et~al.
\newblock Recent advances in convolutional neural networks.
\newblock 2015.

\bibitem[He et~al.(2014)He, Zhang, Ren, and Sun]{he2014spatial}
Kaiming He, Xiangyu Zhang, Shaoqing Ren, and Jian Sun.
\newblock Spatial pyramid pooling in deep convolutional networks for visual
  recognition.
\newblock In \emph{European conference on computer vision}, pages 346--361.
  Springer, 2014.

\bibitem[He et~al.(2017)He, Gkioxari, Doll{\'a}r, and Girshick]{he2017mask}
Kaiming He, Georgia Gkioxari, Piotr Doll{\'a}r, and Ross Girshick.
\newblock Mask r-cnn.
\newblock In \emph{Computer Vision (ICCV), 2017 IEEE International Conference
  on}, pages 2980--2988. IEEE, 2017.

\bibitem[Johnson et~al.(2015)Johnson, Krishna, Stark, Li, Shamma, Bernstein,
  and Fei-Fei]{johnson2015image}
Justin Johnson, Ranjay Krishna, Michael Stark, Li-Jia Li, David Shamma, Michael
  Bernstein, and Li~Fei-Fei.
\newblock Image retrieval using scene graphs.
\newblock In \emph{Proceedings of the IEEE conference on computer vision and
  pattern recognition}, pages 3668--3678, 2015.

\bibitem[Krishna et~al.(2018)Krishna, Chami, Bernstein, and
  Fei-Fei]{krishna2018referring}
Ranjay Krishna, Ines Chami, Michael Bernstein, and Li~Fei-Fei.
\newblock Referring relationships.
\newblock In \emph{Proceedings of the IEEE Conference on Computer Vision and
  Pattern Recognition}, pages 6867--6876, 2018.

\bibitem[Lu et~al.(2016)Lu, Krishna, Bernstein, and Fei-Fei]{lu2016visual}
Cewu Lu, Ranjay Krishna, Michael Bernstein, and Li~Fei-Fei.
\newblock Visual relationship detection with language priors.
\newblock In \emph{European Conference on Computer Vision}, pages 852--869.
  Springer, 2016.

\bibitem[Nguyen et~al.(2016)Nguyen, Yosinski, and
  Clune]{nguyen2016multifaceted}
Anh Nguyen, Jason Yosinski, and Jeff Clune.
\newblock Multifaceted feature visualization: Uncovering the different types of
  features learned by each neuron in deep neural networks.
\newblock 2016.

\bibitem[Ren et~al.(2015)Ren, He, Girshick, and Sun]{ren2015faster}
Shaoqing Ren, Kaiming He, Ross Girshick, and Jian Sun.
\newblock Faster r-cnn: Towards real-time object detection with region proposal
  networks.
\newblock In \emph{Advances in neural information processing systems}, pages
  91--99, 2015.

\bibitem[Russakovsky et~al.(2015)Russakovsky, Deng, Su, Krause, Satheesh, Ma,
  Huang, Karpathy, Khosla, Bernstein, et~al.]{russakovsky2015imagenet}
Olga Russakovsky, Jia Deng, Hao Su, Jonathan Krause, Sanjeev Satheesh, Sean Ma,
  Zhiheng Huang, Andrej Karpathy, Aditya Khosla, Michael Bernstein, et~al.
\newblock Imagenet large scale visual recognition challenge.
\newblock \emph{International Journal of Computer Vision}, 115\penalty0
  (3):\penalty0 211--252, 2015.

\bibitem[Sadeghi and Farhadi(2011)]{sadeghi2011recognition}
Mohammad~Amin Sadeghi and Ali Farhadi.
\newblock Recognition using visual phrases.
\newblock In \emph{Computer Vision and Pattern Recognition (CVPR), 2011 IEEE
  Conference on}, pages 1745--1752. IEEE, 2011.

\bibitem[Sermanet et~al.(2013)Sermanet, Eigen, Zhang, Mathieu, Fergus, and
  LeCun]{sermanet2013overfeat}
Pierre Sermanet, David Eigen, Xiang Zhang, Micha{\"e}l Mathieu, Rob Fergus, and
  Yann LeCun.
\newblock Overfeat: Integrated recognition, localization and detection using
  convolutional networks.
\newblock 2013.

\bibitem[Simonyan et~al.(2013)Simonyan, Vedaldi, and
  Zisserman]{simonyan2013deep}
Karen Simonyan, Andrea Vedaldi, and Andrew Zisserman.
\newblock Deep inside convolutional networks: Visualising image classification
  models and saliency maps.
\newblock 2013.

\bibitem[Urdan(2011)]{urdan2011statistics}
Timothy~C Urdan.
\newblock \emph{Statistics in plain English}.
\newblock Routledge, 2011.

\bibitem[Yosinski et~al.(2015)Yosinski, Clune, Nguyen, Fuchs, and
  Lipson]{yosinski2015understanding}
Jason Yosinski, Jeff Clune, Anh Nguyen, Thomas Fuchs, and Hod Lipson.
\newblock Understanding neural networks through deep visualization.
\newblock 2015.

\bibitem[Yu et~al.(2017)Yu, Li, Morariu, and Davis]{yu2017visual}
Ruichi Yu, Ang Li, Vlad~I Morariu, and Larry~S Davis.
\newblock Visual relationship detection with internal and external linguistic
  knowledge distillation.
\newblock In \emph{IEEE International Conference on Computer Vision (ICCV)},
  2017.

\bibitem[Zhang et~al.(2017)Zhang, Kyaw, Chang, and Chua]{zhang2017visual}
Hanwang Zhang, Zawlin Kyaw, Shih-Fu Chang, and Tat-Seng Chua.
\newblock Visual translation embedding network for visual relation detection.
\newblock In \emph{CVPR}, volume~1, page~5, 2017.

\bibitem[Zhang et~al.(2016)Zhang, Goyal, Summers-Stay, Batra, and
  Parikh]{zhang2016yin}
Peng Zhang, Yash Goyal, Douglas Summers-Stay, Dhruv Batra, and Devi Parikh.
\newblock Yin and yang: Balancing and answering binary visual questions.
\newblock In \emph{Proceedings of the IEEE Conference on Computer Vision and
  Pattern Recognition}, pages 5014--5022, 2016.

\bibitem[Zhou et~al.(2014)Zhou, Khosla, Lapedriza, Oliva, and
  Torralba]{zhou2014object}
Bolei Zhou, Aditya Khosla, Agata Lapedriza, Aude Oliva, and Antonio Torralba.
\newblock Object detectors emerge in deep scene cnns.
\newblock 2014.

\bibitem[Zhou et~al.(2016)Zhou, Khosla, Lapedriza, Oliva, and
  Torralba]{zhou2016learning}
Bolei Zhou, Aditya Khosla, Agata Lapedriza, Aude Oliva, and Antonio Torralba.
\newblock Learning deep features for discriminative localization.
\newblock In \emph{Proceedings of the IEEE Conference on Computer Vision and
  Pattern Recognition}, pages 2921--2929, 2016.

\bibitem[Zhu and Jiang(2018)]{zhu2018deep}
Yaohui Zhu and Shuqiang Jiang.
\newblock Deep structured learning for visual relationship detection.
\newblock In \emph{Proceedings of the Thirty-Second {AAAI} Conference on
  Artificial Intelligence, New Orleans, Louisiana, USA, February 2-7, 2018},
  2018.

\end{thebibliography}
	
	\appendix
	
	\section{Use Cases}
	\label{sec:app_usecases}
	This section explains how the method proposed in this paper may be used for visual relationship detection, and other areas.
	\\
	
\begin{figure}[h]
		\centering
		\includegraphics[width=\linewidth]{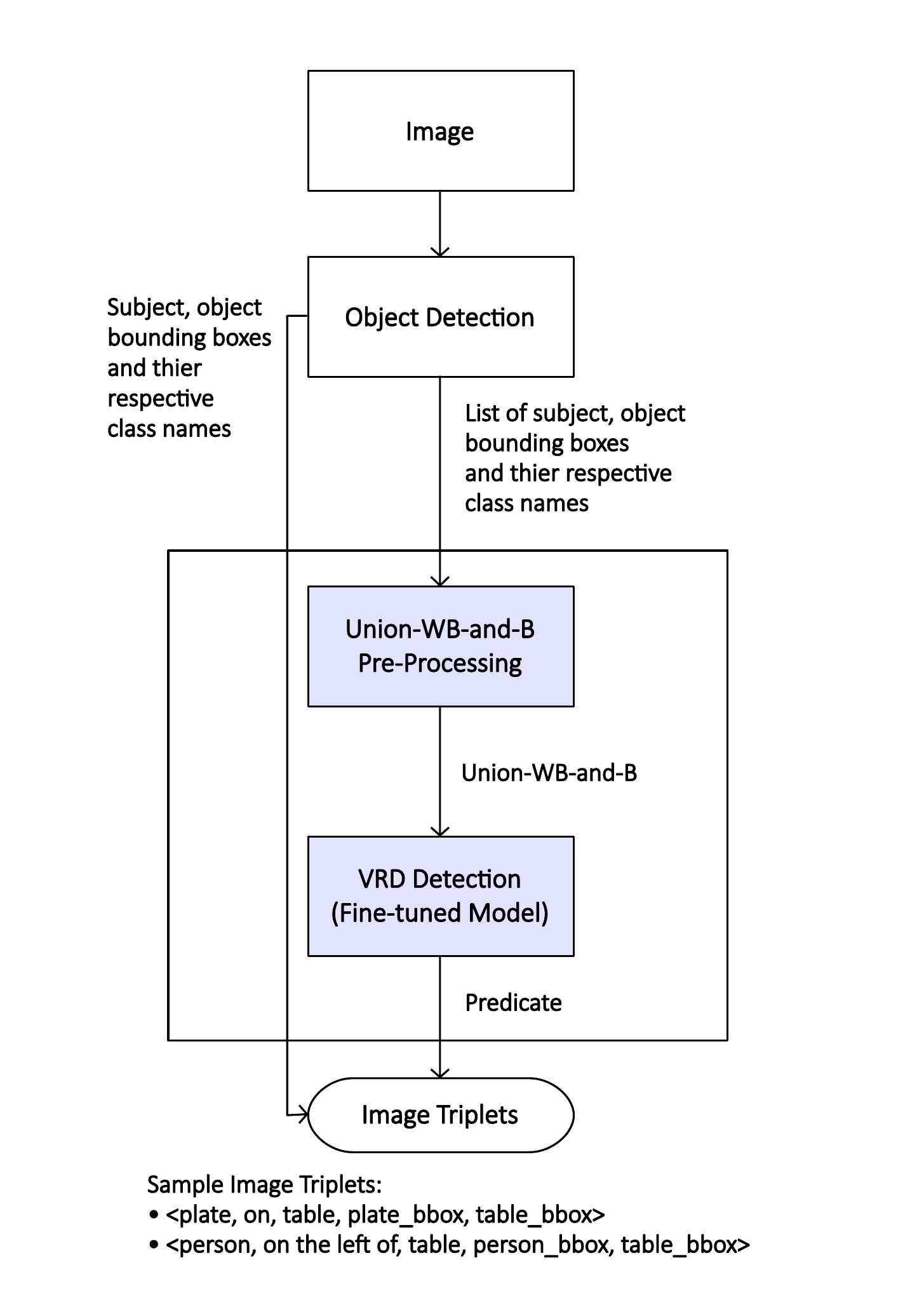}
		\caption{The general structure -- depicting how the fine-tuned models may be used to generate the triplets from a given image.}
		\label{fig:uc_acg}
	\end{figure}

	Fig. \ref{fig:uc_acg} depicts\footnote{The blue boxes depict the research carried out in this paper.} how an object detector such as \ac{fasterrcnn} and the fine-tuned model (\ac{eg} Union-WB-and-B) may be used to predict the triplets. This is achieved by first feeding the respective image into the object detector. The detected objects bounding boxes are then used to generate pairs of objects which are the used to generate the Union-WB-and-B pre-processed images. These are then fed into the respective fine-tuned model to predict the respective predicate. In the final stage, the triplets are generated using the objects bounding boxes, class names (provided by the object detector) and the respective predicate (provided by the fine-tuned model). Therefore, resulting in a triplet for each pair of detected objects.
	\\
	
	 The predicted triplets provide the basis for other computer vision problems such as automatic caption generation, image retrieval, \ac{vqa}, and accessibility\footnote{To assist both blind and visual impaired persons to understand the world around them and carry out the desired task}. The following is a brief explanation of how the predicted triplets (as depicted in Fig. \ref{fig:uc_acg}) may be used to achieve these tasks:
	 \begin{itemize}
	 	\item \textbf{Automatic Caption Generation} --- use the triplets to retrieve sentences from the internet that are relevant to the respective image (\ac{ie} explain the image content);
	 	\item \textbf{Image Retrieval} --- index the triplets and use them when searching for an image based on the image content;
	 	\item \textbf{\ac{vqa}} --- use an \ac{lstm} neural network to understand the question and then use the image triplets to answer the question; and
	 	\item \textbf{Accessibility} --- for each video stream frame, generate the respective triplets, generate sentences and then use text-to-speech to inform the user accordingly.
	\end{itemize}
	
\end{document}